%% file: main.tex
\documentclass[letterpaper, 10 pt, conference]{ieeeconf}  
\IEEEoverridecommandlockouts
\usepackage{graphicx}
\usepackage{mathptmx} % assumes new font selection scheme installed
\usepackage{times} % assumes new font selection scheme installed
\usepackage{amsmath} % assumes amsmath package installed
\usepackage{amssymb}  % assumes amsmath package installed
\usepackage{units}
\usepackage{verbatim}  % multiline comments via \begin{comment}\end{comment}
\usepackage{nicefrac}		% to get better units
\usepackage{acronym}		% expand first occurance of acronyms
\usepackage{glossaries}		% acronyms + glossaries
\usepackage[usenames,dvipsnames]{xcolor}			% extened color package
\usepackage{cite} % smarter citations, e.g.\ makes [1][2][3] into [1]-[3]
\usepackage{url} % \url{my_url_here}
\usepackage[multidot]{grffile} % allow dots in graphics filenames
\usepackage{multirow}
\usepackage{booktabs}
\makeatletter
\let\NAT@parse\undefined
\makeatother
\usepackage{hyperref}
\hypersetup{colorlinks,allcolors=black}
\makeglossaries

\input{MathMacros} % for iso-standard math
\input{Acronyms}

\title{\LARGE \bf The Need for Inherently Privacy-Preserving Vision\\in Trustworthy Autonomous Systems }

\author{%
Adam K.~Taras$^{1}$,
Niko Suenderhauf$^{2}$, 
Peter Corke$^{2}$, and
Donald G.~Dansereau$^{1}$
\thanks{$^{1}$Australian Centre For Robotics, School of Aerospace, Mechanical and Mechatronic Engineering, The University of Sydney.
{\tt\small adam.taras, donald.dansereau@sydney.edu.au}}%
\thanks{$^{2}$Queensland University of Technology (QUT) Centre for Robotics, Brisbane, Australia. {\tt\small niko.suenderhauf, peter.corke@qut.edu.au}}%
\thanks{This work was partially supported by the QUT Centre for Robotics.}
}

\begin{document}

\maketitle
\thispagestyle{empty}
\pagestyle{empty}

%%%%%%%%%%%%%%%%%%%%%%%%%%%%%%%%%%%%%%%%%%%%%%%%%%%%%%%%%%%%%%%%%%%%%%%%%%%%%%%%

\begin{abstract}
Vision is a popular and effective sensor for robotics from which we can derive rich information about the environment: the geometry and semantics of the scene, as well as the age, gender, identity, activity and even emotional state of humans within that scene.  This raises important questions about the reach, lifespan, and potential misuse of this information.  This paper is a call to action to consider privacy in the context of robotic vision. We propose a specific form privacy preservation in which no images are captured or could be reconstructed by an attacker even with full remote access. We present a set of principles by which such systems can be designed, and through a case study in localisation demonstrate in simulation a specific implementation that delivers an important robotic capability in an inherently privacy-preserving manner. This is a first step, and we hope to inspire future works that expand the range of applications open to sighted robotic systems.
\end{abstract}

%%%%%%%%%%%%%%%%%%%%%%%%%%%%%%%%%%%%%%%%%%%%%%%%%%%%%%%%%%%%%%%%%%%%%%%%%%%%%%%%

%===============================================================================
\section{Introduction}
\label{sec:intro}

% Example citation~\cite{hinojosa2021learning}. Example acronym \gls{SLAM}, like \gls{SLAM}.

% Here we introduce the paper in broad strokes, establish motivation (who cares?)

Do you have a robot vacuum cleaner? Perhaps one of the new generation robots that uses a camera to navigate around your house? Where do those camera images go? Who can see them? Perhaps the images should never leave your house. Perhaps they should never leave the robot or the camera chip. Perhaps, to best protect your privacy, the images, as we know them, should never be formed in the first place. 

Research in robotic vision has neglected and often ignored the legitimate privacy concerns of potential end-users, and instead focused solely on improving task performance~\cite{eick2020enhancing}. 
We argue that this short-sighted strategy ultimately forestalls the widespread adoption and societal impact of robotic vision. In contrast, we propose to re-imagine robotic vision to achieve an optimal balance between task performance and privacy protection.

This paper is a call to action for the robotic vision community to develop novel computational imaging technology for privacy-preserving robotic vision. By developing novel combinations of optical, analogue, and algorithmic elements, the community -- academia and industry -- could build novel camera technology that never forms human-interpretable conventional images, and from which such images could never be reconstructed from the sensor data.

Such new camera designs would address the legitimate privacy concerns that are impeding the beneficial adoption of robotics in applications of societal and economic importance, e.g.\ where there is a strong emphasis on social human-robot interactions (healthcare, aged care); where robots and humans collaborate and intellectual property must be protected (manufacturing); or where a breach of privacy could have safety and security implications (energy) or impede sovereign capabilities (defence). By addressing these legitimate privacy concerns, novel privacy-preserving camera technology will broaden the applicability, and increase the public acceptance, of robotic vision applied to these domains without compromising the privacy and security of citizens, industries and governments. 

Our paper first introduces privacy as a concept in the context of robotic vision (\S \ref{sec:privacy}) and discusses current approaches to privacy preserving robotic vision (\S \ref{sec:prior_work}). We then introduce our proposed concept for inherently privacy-preserving vision systems (\S \ref{sec:approach}) and present a localisation case study that exemplifies this approach (\S \ref{sec:localisation}). We close with a discussion (\S \ref{sec:discussion}) and hope to elicit input and opinions from the research community.

\section{What is Privacy?}
\label{sec:privacy}
Privacy is a complex concept that is relevant to many areas of society. Interestingly, it was the increased availability of easy-to-use photography cameras that motivated the definition of privacy as ``the right to be let alone'' in an 1890 law review article~\cite{brandeis1890right}.

Since then, numerous definitions and analyses of privacy have been published, with Altman’s ``selective control of access to the self or to one’s group''~\cite{altman1976privacy} one of the most prominent. As reviewed in~\cite{leino2001privacy}, the current research literature distinguishes physical, psychological, social, and informational privacy. 
These respectively relate to concepts such as personal space or physical access; the right to control with whom and under what circumstances to share one's thoughts; the ability to control anonymity and social interactions; and when, how and to what extent information about the self will be released to another person or organisation~\cite{leino2001privacy}.

Although all forms of privacy can potentially be violated by robots and are therefore relevant to the study of robotics, our project focuses on a specific form of informational privacy. 
Concretely, in this paper, we understand \emph{privacy preservation} to be the minimisation of the risk of exposing a human-interpretable image of the environment in which a robotic vision system operates, or the risk of exposing information that enables the reconstruction of such an image.

The robotics community largely considers privacy concerns and task performance to be orthogonal issues: of 89,120 papers published in the top robotics journals and conferences 1982-2019, only 0.5\% mention privacy~\cite{eick2020enhancing}, despite the fact that at least 132 countries now have data privacy laws, and data protection officials from over 60 countries have expressed concerns about the impacts of robotics and AI on privacy~\cite{eick2020enhancing}. The disconnect between the robotics research community and these recent developments is a clear call to action, and our motivation.

%===============================================================================
\section{Current Approaches to Privacy in Vision}
\label{sec:prior_work}

\begin{figure*}[ht!]
	\centering
    \includegraphics[width=1\linewidth]{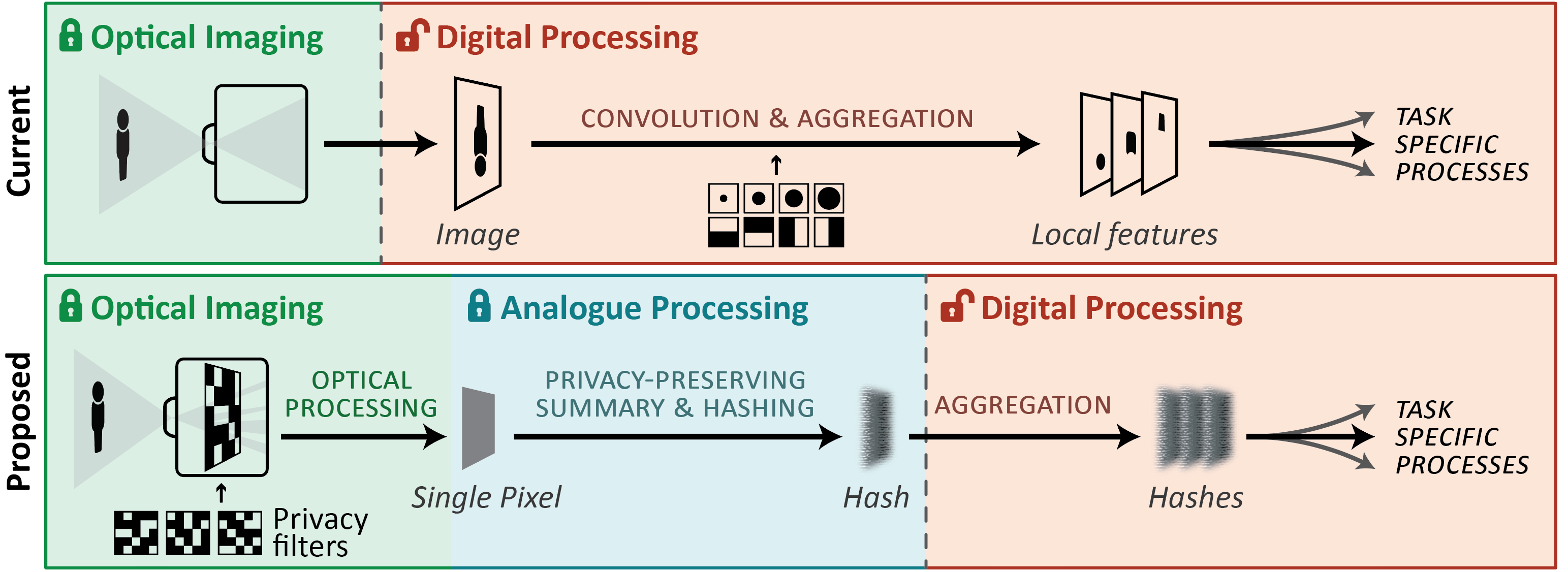} 
	\caption{Example of the proposed inherently privacy-preserving approach to vision: Current robotic vision (top) uses conventional optics and cameras to form human-interpretable images with many 100,000s of pixels. Local features are often extracted through a series of learned convolutions and aggregations, e.g.\ by convolutional neural networks, before being processed by task-specific components, e.g.\ for object detection or grasp synthesis. 
    Instead, we propose to shift processing into the optical-analogue domain (bottom): in this example a micromirror device implements a series of learned filters through which the light in a scene enters a single-pixel sensor. Analogue processing performs privacy-preserving summarisation and hashing \emph{before} the data enters the digital domain, where it is vulnerable to attacks and security breaches. Specialised task-specific algorithms and learned components operate on the secure hashes in the digital domain to perform important robotics tasks. Privacy is ensured because reverting the optical-analogue hashing is intractable.} 
    \label{fig:convPipelineCompare} 
\end{figure*}

Even among those tackling the specific task of privacy-preserving vision, there are multiple definitions of privacy in use. For example, Zhang et al.~\cite{zhang2012privacy} propose the use of \gls{RGB-D} cameras for masking foreground objects and using only depth information to maintain user privacy. Many would also argue that capturing of depth-only imagery is an unacceptable breach of privacy. 

Much of the current work for privacy preserving vision seeks greater levels of obfuscation or even encryption. These works fall broadly under two categories, depending on when the obscuration occurs: after capturing digital images, or during the image formation process. Here by obfuscation we mean representing signals in a difficult-to-interpret form, and by encryption we entail encoding with a cryptographic key such that information can only be compromised if someone knows or deduces the key. 

In this section, we discuss encouraging results spanning these approaches, and identify key gaps that represent the opportunity at the heart of this paper.

The fundamental law of information recovery~\cite{dwork2014algorithmic} states that ``overly accurate answers to too many questions will destroy privacy in a spectacular way''. We use this as a guiding principle in understanding current approaches to privacy preservation, employing the most conservative position that anything that is digitised should be suspect as open to unwelcome observers. While vision systems may summarise images with sparse representations, e.g.\ through feature extraction, Pittaluga et al.~\cite{pittaluga2019revealing} show that even sparse information like \gls{3D} point clouds of \gls{SIFT} features can be used to reconstruct colour images of the complete scene. This reconstruction succeeds even after removing keypoint orientation and scale. 

Reconstruction from features is a strong cautionary result. Even if feature extraction is implemented before digitisation it is not guaranteed to be privacy preserving, but rather the features themselves must be robust against reconstruction and various attacks. Thus, key characteristics used to judge the privacy-preserving capabilities of systems are the amount of information captured as well as by what means information is either discarded, obfuscated, or encrypted.

\subsection{Post-capture Obfuscation and Encryption}

Methods in this category capture digital images then extract and obscure key information before discarding the images. Any system that captures digital images is open to attack via unsecured remote access to the compute system. Of even greater concern is that in some cases the obscured form of the images that are generally taken to be privacy-preserving nonetheless provide enough information to reconstruct imagery of the scene. 

A concrete example of the later arises in the random feature lines proposed by Speciale et al.~\cite{speciale2019privacy, speciale2019privacy2}. This work summarises scenes in terms of obfuscated keypoints, replacing~\gls{SIFT} features with randomly oriented lines in three dimensions. This obfuscates the contents of the feature cloud, rendering scenes unrecognisable to the human eye. However, Chelani et al.~\cite{chelani2021privacy} later found that images can be reconstructed from these obscured feature clouds by exploiting statistics of nearby feature points. This cautionary example indicates that there is a wealth of structure and redundancy in even obscured representations, and points to the need for a more significant re-think of how we carry out vision to preserve privacy.

Conventional encryption~\cite{kaur2020comprehensive}, whilst relatively inexpensive and broadly applicable, is also subject to breaches. Encrypted data may be open to unauthorised access, and data breaches due to human error are uncomfortably common and increasing in frequency~\cite{ayyagari2012exploratory}. Ultimately, encrypted data stored in the cloud~\cite{subashini2011survey} is only as secure as the agents entrusted with the decryption keys. This opens the potential for spoofing attacks, requires securely distributing keys, or requires algorithms that work on encrypted imagery without requiring decryption, all of which are challenging.

\subsection{Optical Obfuscation and Encryption}

There are several notable examples where the camera is involved in the process of obfuscating or encrypting imagery. Key-Nets~\cite{byrne2020key} use custom optical fibre bundles and custom imaging sensors with per-pixel bias and gain to effectively carry out a Hill Cipher~\cite{hill1931concerning}. The paper proposes a way of converting neural architectures to operate on the encoded imagery, offering the potential for existing vision pipelines to be used on keyed data. However, this approach requires custom optical fibre bundles to shuffle pixels and custom silicon  implementing per-pixel bias and gain. Manufacture is thus extremely impractical, preventing widespread adoption of this approach. It is also the case that this approach, like other forms of cryptography, is only as secure as storage of a private key, and is open to a variety of attacks.

There is growing interest in the use of optical neural networks including diffractive deep neural networks~\cite{lin2018all, bai2022image}. Recent work shows that diffractive layers can be designed such that destructive intereference occurs for all except a target class. The camera captures data with low latency as all computation is optical, and manufacture of printed masks is practical and broadly accessible. A key limitation is that the proposed diffractive cameras require narrowband, collimated active illumination to function, limiting their widespread deployment. Hard guarantees about leaking of private information through the diffractive imaging process are also unclear.

Finally, there is growing interest in reconstruction-free vision systems, in which encoded imagery is captured but never converted to a human-interpretable form. An example of this is the use of a lensless imaging for action recognition~\cite{wang2019privacy}. Whilst capturing obfuscated imagery, this work does not explicitly address the potential for an attacker to reconstruct human-interpretable imagery. Based on recent progress in lensless imaging that explicitly reconstructs human-interpretable imagery~\cite{boominathan2022recent}, it seems inevitable that some form of reconstruction should be possible in these cameras.  Another example of these systems is image classification from single-pixel cameras~\cite{latorre2019online}. Such systems are likely vulnerable to reconstruction as evidenced by techniques in compressive sensing.

\subsection{Key Gaps}
It is evident that prior works either digitise signals and then make them private, and are open to digital attack; or involve the camera and do not prevent reconstruction. A single exception, Key-nets, is impractical due to the requirement for custom optical and silicon manufacture. 

In the following we propose an approach to inherently privacy-preserving vision that involves imaging the scene through optical-analogue computation such that only secure, privacy-preserving information is ever transferred into the digital domain. Relative to prior work, key differences are in the nature and quantity of digital information, and that reconstruction should not be possible. Our approach is extensible so that depending on application, other definitions of privacy can be engineered into the camera. For example, in some cases making an inference of the existence of an object in the scene could be a violation of privacy. We believe this represents a unique approach to privacy preservation, opening new application areas where systems with vision cannot operate at present.

By judiciously moving computation out of the digital domain, removing and summarising information, and obfuscating or encrypting the information that remains in the optical-analogue domain, our proposed approach offers a different class of privacy-preserving camera that we call inherently private vision systems. 

%===============================================================================
\section{Inherently Privacy-Preserving Vision}
\label{sec:approach}
In this work we propose the concept of an inherently privacy-preserving vision system. This is one in which none of the attacks outlined in the previous section could be applied: brute-force decryption, image decoding, spoofing attacks and data breaches, and access to digital imagery through unauthorised remote access to a robot's hardware should not be possible. Privacy in this sense means that at no point in the system are digital images stored, nor could they be reconstructed. 

Inspired by work on custom optics and sensors~\cite{byrne2020key} as well as optical neural networks~\cite{bai2022image}, we propose that inherently private vision is possible by constructing custom cameras designed to carry out specific tasks chiefly through their optics and analogue electronics. These cameras must be designed following a set of principles that prevent the digitisation of private information. 

We propose here a starting point for this set of ideals for constructing inherently privacy-preserving vision systems:

\begin{itemize}
    \item Specialise the camera to the task; this sacrifices generality for privacy as the camera can only be used for the task(s) it's designed for,
    \item Shift as much processing as possible out of the digital domain, keeping it out of reach of remote attack,
    \item Maximise information-destroying operations prior to digitisation,
    \item Apply obscuration prior to digitisation such that brute-force attack becomes the only option for inverting the imaging process, 
    \item Consider all information already available to the attacker, e.g.\ sequences of data and priors to improve domain performance and ensure privacy, and
    \item Maximise ambiguity, so that even a successful brute-force inversion of the imaging process is not likely to yield the correct image.
\end{itemize}

We anticipate a broad variety of implementations could meet the above principles, and in this paper we conduct a case study for carrying out localisation. Comparison of conventional imaging and our specific implementation employing optical-analogue single-pixel hashing are depicted in Fig.~\ref{fig:convPipelineCompare}, and a detailed proof-of-concept study carried out in simulation is included in the following section. In this implementation we address the principles laid out above by specialising a camera to the task of localisation, shifting much of the machinery of localisation into the optics and analogue electronics of the camera, and employing information-destroying feature extraction, summarisation, and hashing prior to digitisation such than many images yield identical hashes that are nevertheless useful for localisation.

\section{Case Study: Privacy-Preserving Localisation}
\label{sec:localisation}

Here we apply the principles laid out in the previous section for designing inherently privacy-preserving vision systems for the robotic task of localisation. We show in simulation that by shifting digital processing into optics and analogue electronics, we can accomplish effective localisation without ever capturing digital images and without capturing enough information to allow reconstruction of these images. 

The overall approach of this study is to build on the architecture of a single-pixel camera to carry out feature extraction and summarisation in the optical-analogue domain. In this architecture a series of masks are applied to a wide-field single pixel, and the resulting signal passed through bespoke analogue computation. Applying insights from existing feature-based methods, we select masks and subsequent summarisation that represents a sort of hashing or fingerprinting, such that information is destroyed before digitisation. We evaluate the proposed approach by solving an image retrieval problem analogous to localisation, demonstrating accuracy on par with a standard \gls{SIFT}-based approach.

\subsection{Why Localisation?}

A robotic vacuum cleaner working in medical settings, a warehouse cart handling sensitive intellectual property in manufacturing and a drone delivering goods over government buildings all rely on an understating of position within the scene. With current vision systems, however, any data collected would be at risk to attack. By addressing the problem of privacy-preserving localisation, we address privacy issues for a breadth of application domains. We anticipate that solving the localisation problem can also give direct insight into more complex vision tasks such as object tracking or grasping. 

Localisation tasks range from image retrieval~\cite{ledwich2004reduced,chaari2007global}, place recognition~\cite{garcia2017hierarchical}, or full 6 degree of freedom pose regression~\cite{sattler2019understanding, ott2020vipr}. Here we address the image retrieval problem for privacy preserving localisation. 

\subsection{Optical-Analogue Image Hashing}

We first identify and design methods of image retrieval such that the signal processing in the optical-analogue domain computes image hashes with high utility but limited private information. The hash function must be tractable in hardware, information-destroying, and descriptive enough to allow localisation.
We note that intuitively local hashes may be more open to exploitation for reconstruction as they reveal local structure, while global hashes introduce more ambiguity. Thus, we restrict our pipeline to using global features.

\begin{figure}
    \centering
    \includegraphics[width=0.75\columnwidth]{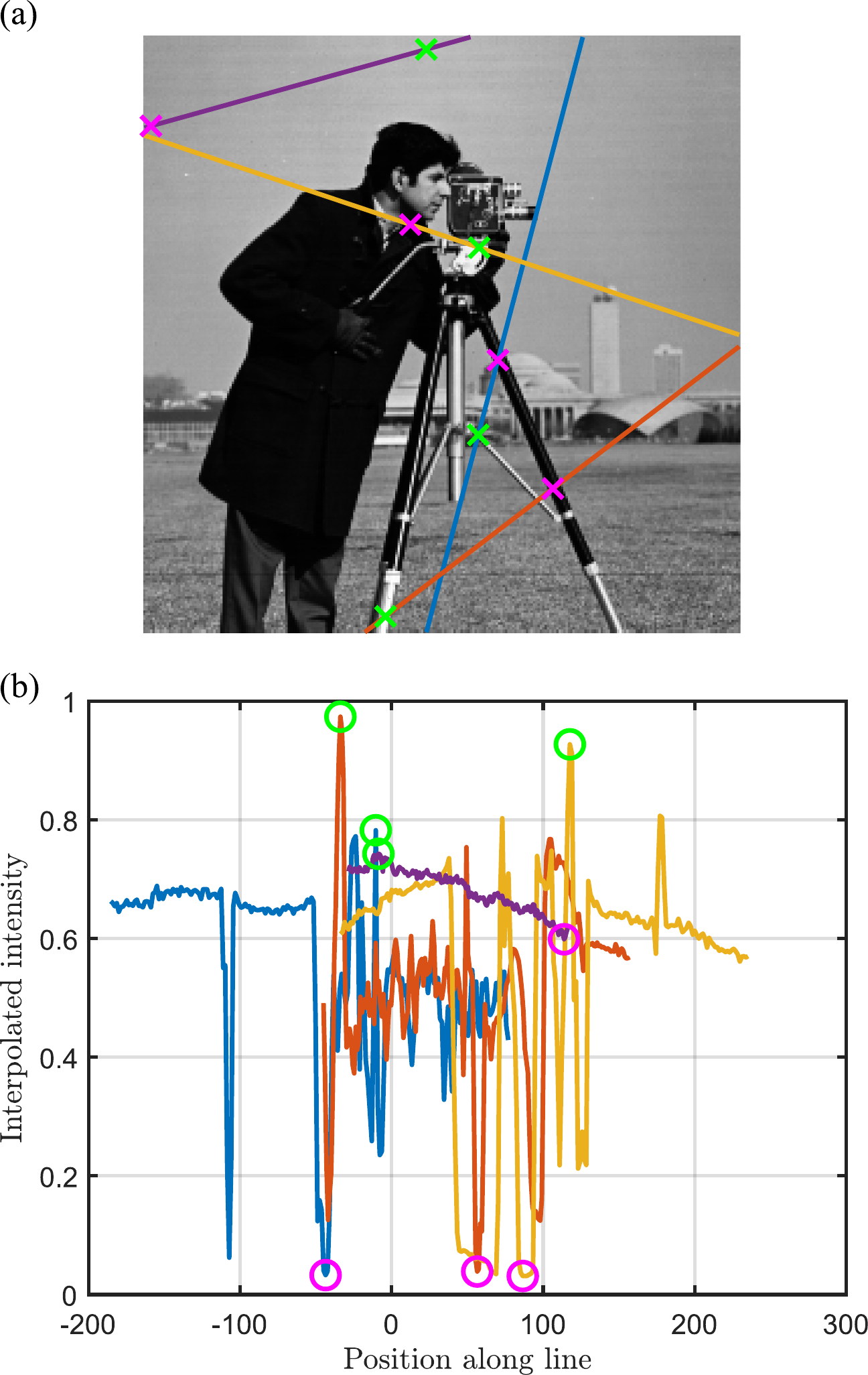}
    \caption{Optical-analogue summarisation for localisation: (a)~an input scene is sampled along four randomly generated lines, recording only the  global maxima (green) and minima (magenta) along each line, (b)~traces along each line with extrema highlighted. The hashing process accumulates these extrema prior to digitisation, destroying and obscuring information about the scene.}
    \label{fig:cameramanLineTraces}
\end{figure}

Edges play an important role in image understanding, and in the context of our single-pixel architecture a simple way of looking for edges is to incrementally project masks that admit light along lines. We measure extrema along the masked lines by using maximum and minimum hold circuits. We accumulate the resulting pairs of maxima and minima, one per line, over $N$ lines. For randomly selected and ordered lines, the resulting accumulation of pairs of extrema reveals little about the structure of the image, while representing a fingerprint that can be used to discern the image from a sequence.

We illustrate the process of measuring this hash for $N=4$ features in \autoref{fig:cameramanLineTraces}, where each feature is the tuple of maximum and minimum along the curve. To look for more edges we measure along more lines, and by randomising the locations of these lines we destroy information about the original structure of the scene while collecting a fingerprint of its content. Note that this process is rotation-invariant for large $N$. 

\begin{figure}
    \centering    
    \includegraphics[width=0.9\columnwidth]{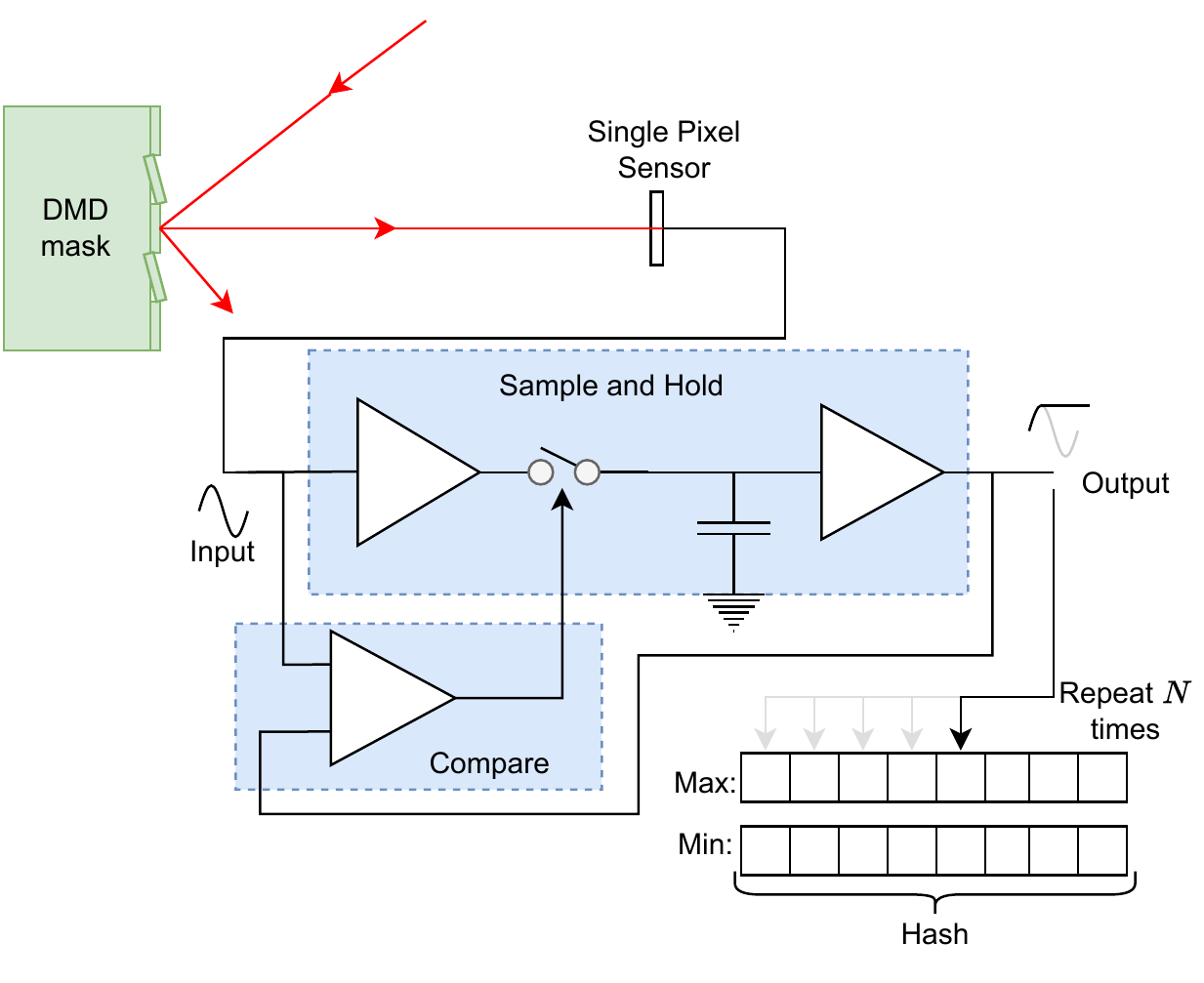}
    \caption{Example hardware implementation for processing in the optical (green) and analogue (blue) domains: A \gls{DMD} applies a set of fixed filters to light from the scene. The signal is sampled by a single pixel and traverses analogue computation that detects extrema over the series of filters. This is repeated over $N$ features, accumulating a histogram or hash-like fingerprint of the scene that is then digitised and loaded to perform localisation. See Fig.~\ref{fig:fingerprintsOfDiffScenes} for example hashes.}
    \label{fig:analogue_setup}
\end{figure}

We depict a hardware implementation of the line hash in \autoref{fig:analogue_setup}. To maintain privacy the \gls{DMD} is driven through a fixed set patterns that cannot be changed. This limits flexibility of the camera, but is critical to prevent outside attack.

Variations in the hashing process are possible by altering the masks employed by the \gls{DMD}. One concern with the random lines is that they begin and end at the boundaries of the images, and under motion this could allow an attacker to infer details of the image boundaries. This motivates an alternative hash which computes extrema over circles rather than lines. Selecting random radii and positions yields similar properties to the random-line approach, including rotation invariance. For input images of dimension $1280\times 720$, radii drawn from the uniform distribution of $[15,50]$ pixels showed strong results.

\begin{figure}
    \centering
    \includegraphics[width=1\columnwidth]{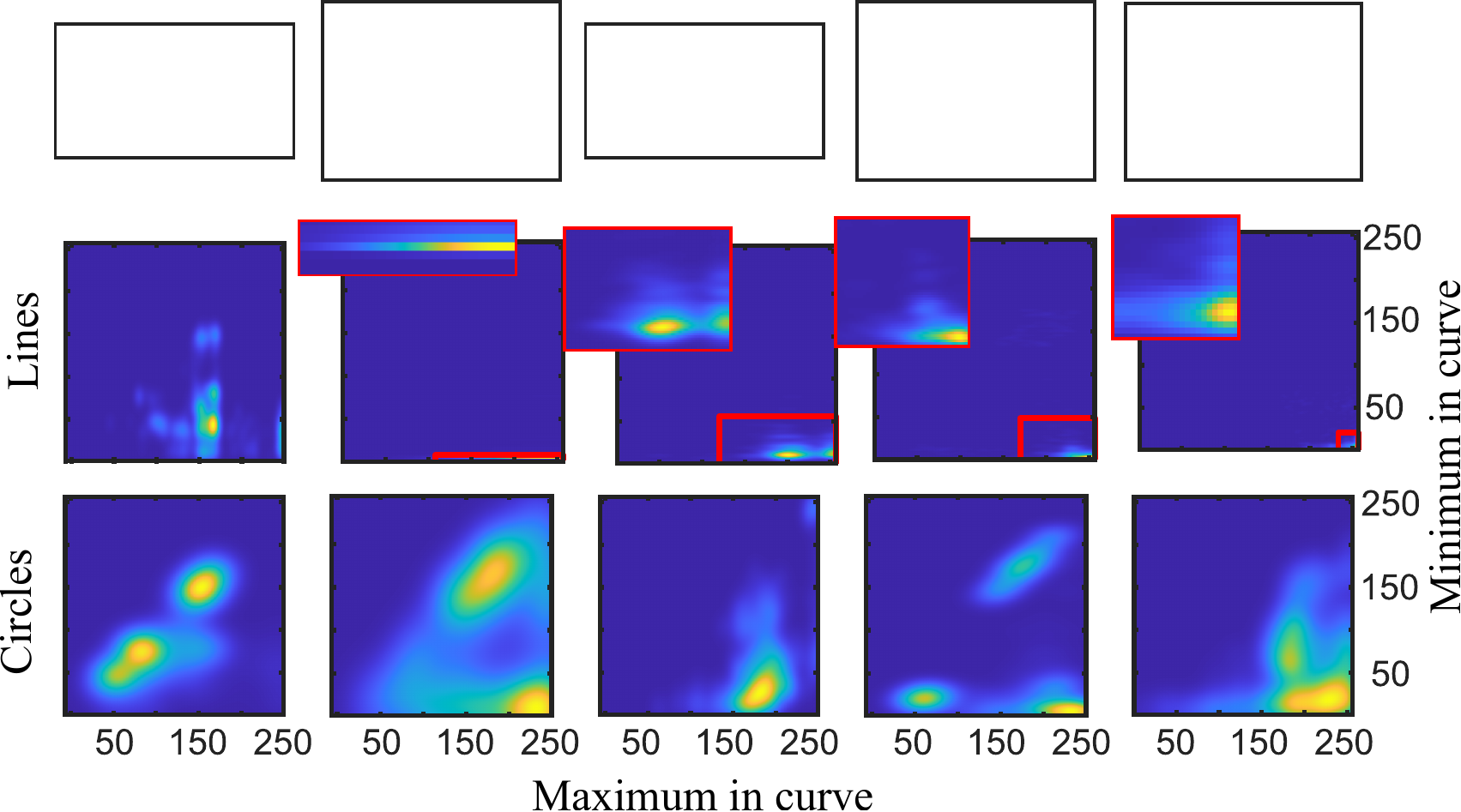} 
    \caption[Comparison of global feature fingerprints for different scene types]{Comparison of global feature fingerprints for different scene types, with each column showing a different location. $10^3$ random line extrema features (middle row) and random circle extrema features (bottom row) with radii sampled from $[15,30]$ pixels show that each image has a unique fingerprint. Insets depict greater detail around features with saturation.}
    \label{fig:fingerprintsOfDiffScenes}
\end{figure}

\autoref{fig:fingerprintsOfDiffScenes} visualises the proposed hashes by plotting histograms of extrema pairs in 2D. These are shown for a range of input scenes. Here we compute hashes over $N=10^3$ random lines or circles, 
and smooth the display for visualisation using kernel density estimation~\cite{peter1985kernel,silverman2018density}. By construction, this visualisation of the hashes must lie below the diagonal. For the random line extrema, each line measures the dominant edge over the dimension of the image. Since this spans the whole image, the features present are strongly affected by the amount of saturation in the image. On the other hand, the random circle extrema are more sensitive to local edges and features. When these features lie close to the diagonal, the scene has textureless regions. In the following we show that these hashes represent fingerprints that are sufficiently unique to allow localisation.

\subsection{Localisation}

To localise based on the proposed hashes, we train a \gls{BoW}-based approach on a dense trajectory of hashes. At inference time, a query hash from the sensor is presented for search and the most similar hash in the reference trajectory is retrieved, localising the robot to the corresponding point in the trajectory. This approach supports a variety of types of visual words, allowing us to directly compare our hashing approach with more conventional, privacy revealing features. We choose \gls{BoW} over neural network-based approaches because, although they may offer superior results, they also show more complex behaviours that can be more difficult to interpret. That our approach works well even with the simpler \gls{BoW} localisation is more revealing.

\subsection{Results}

We evaluate how well \gls{BoW}-based image retrieval is able to predict the position of unseen test images. We use the ``Digiteo Seq 2'' dataset~\cite{el2021indoor}, which contains handheld photos from an office floor. We use a single camera for the trajectory from this stereo dataset. The \gls{BoW} is trained on hashes from a training stride of one in every 20 images, and the remaining images are used for testing. When querying a test image, we consider the localisation to be correct if the true image index is within 30 frames (i.e.\ 1.5 times the training stride) of the \gls{BoW} best-match image.

\begin{figure}
    \centering	\includegraphics[width=0.8\columnwidth]{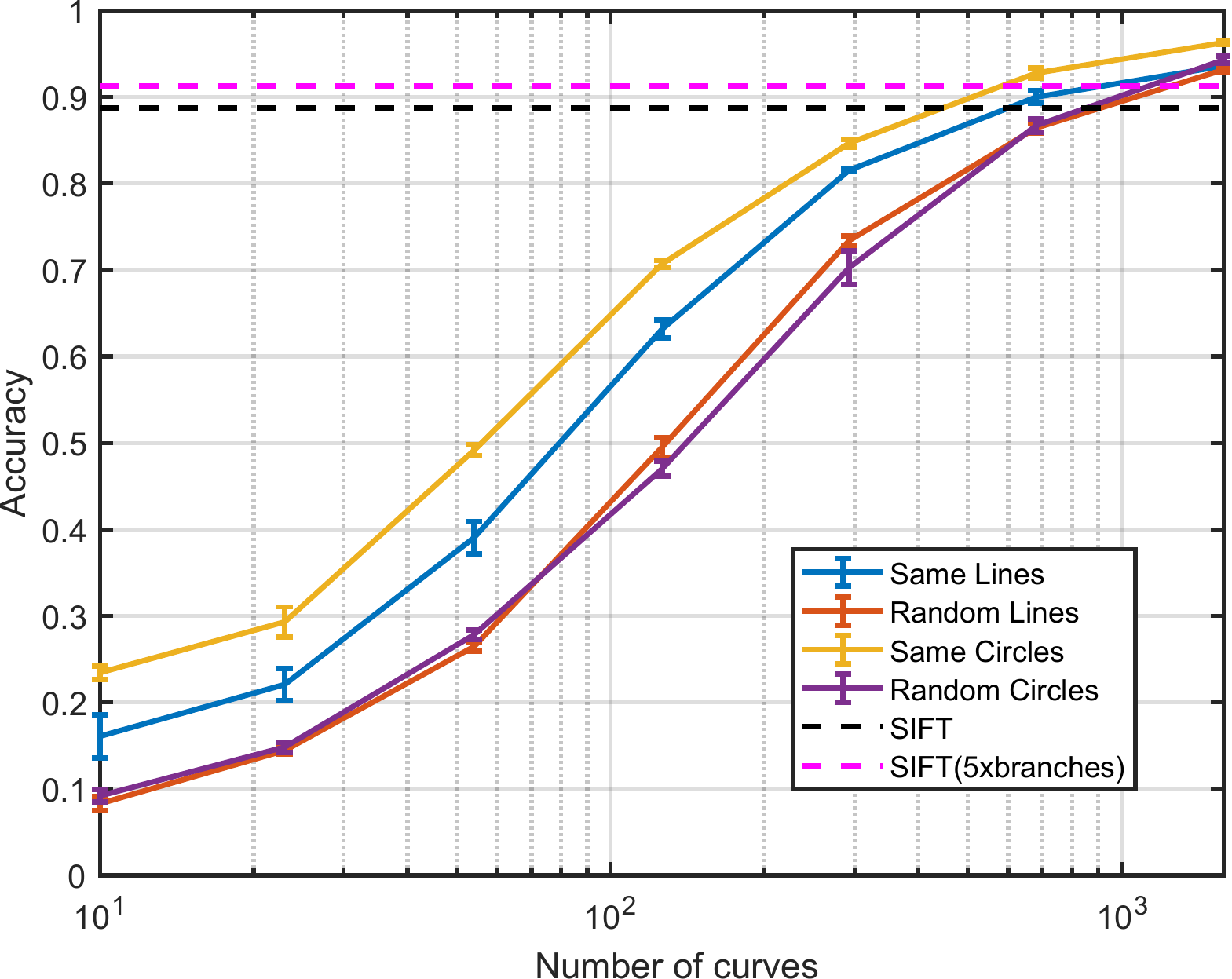} 
	\caption{Accuracy of localisation as a function of the number of features accumulated $N$, and for randomly changing curves or repetition of the same curves for each input scene. The \gls{SIFT} localisation approach is also shown for comparison. While initially performing worse, for large $N$ using randomly changing curves converges to the performance of fixed curves. There is no significant difference between the circle and line methods, and performance exceeds that of the conventional \gls{SIFT}-based approach.} 
 \label{fig:varyNumLines} 
\end{figure}

In \autoref{fig:varyNumLines}, we measure the accuracy of localisation while varying the number of curves used in our approach and compare against the same BoW approach trained on \gls{SIFT} features. For few features performance is weak, but it increases as the number of features is increased. We also consider randomizing the feature curves, circles or lines, for each input image, or using a fixed set. The difference between these is not significant, and it decreases as the number of features increase. 

The proposed methods are ultimately able to meet and slightly exceed the baseline performance of SIFT, with no significant advantage to either lines or circles as curves in this case. The slight variation in SIFT performance indicates that fine tuning of the image retrieval system is useful in optimising performance for a particular dataset or context. We do not claim that these methods outperform \gls{SIFT}, but rather that they are comparable in localisation accuracy while inherently maintaining privacy.

\begin{figure}
    \centering
    \includegraphics[width=1\columnwidth]{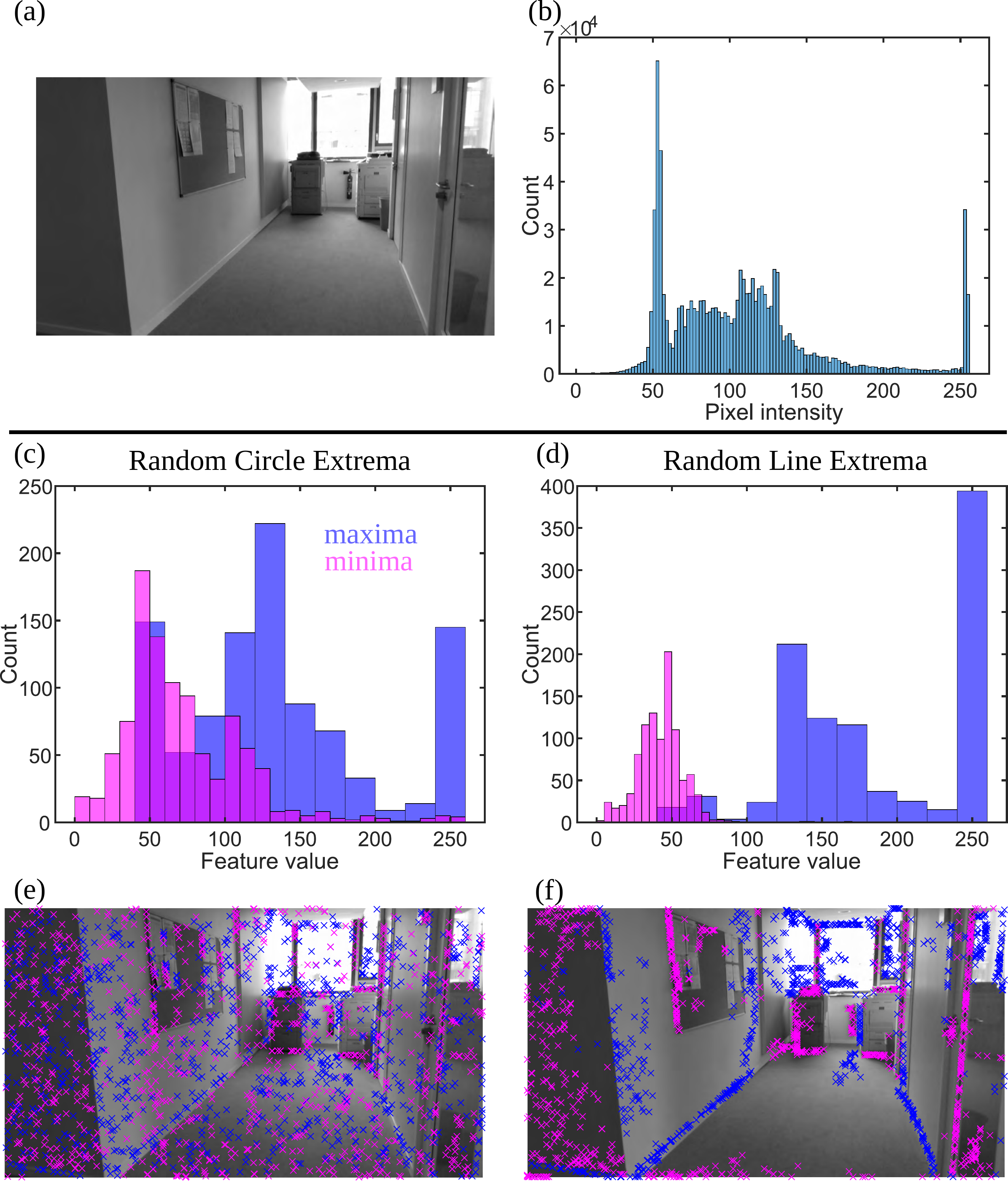}
    \caption{Distribution of sources of data for proposed hashes. (a) original scene (b) distribution of pixel intensities in the scene. Once processed by our random circle extrema pipeline (c,e) or random line extrema (d,f) the distributions do not sample the true distribution evenly. The visualisation of the source of data (e,f) indicates that there are large sparse regions in either case, with lines sampling more densely around extreme regions as expected. The true locations of the data in the hash is not exposed, so an attacker would have to recreate the image without knowing the location of the extrema, only their intensities.}
    \label{fig:location_and_distribution_of_info_comp}
\end{figure}

Empirically we observe that $N\sim10^3$ curves are required for good performance at this task. This is a factor of $10^3$ fewer than the number of pixels in the input (megapixel) images. We investigate which parts of the input images are reflected in the proposed hashes in \autoref{fig:location_and_distribution_of_info_comp}. We see that the hashing procedure both reduces the amount of image information represented and obscures it in keeping with the recommendations in the previous section. The hashing only reveals a small subset of the image, the hashing process hides the locations of the extrema, and the distributions of maxima and minima do not directly reflect the intensity distributions present in the images.  

From these observations we conclude that it is doubtful any algorithm could reconstruct an image from its hash. However, even if one did succeed at this, the fact that a very large number of distinct images produce the same hash would prevent the attacker from knowing if they had constructed the correct image.

%================================================================================
\section{Discussion and Call to Action}
\label{sec:discussion}
 
We proposed a new class of inherently privacy-preserving vision system that fills important gaps in current approaches and opens opportunities for follow-on work. We described a set of principles by which such systems can be designed, by moving processing out of the digital domain, and thus out of reach of remote attackers. The proposed systems never capture images nor do they capture enough information to allow reconstruction of private images. 

We demonstrated our approach through a case study in inherently privacy-preserving localisation. The success of this study in delivering an important robotic capability lends support to the practicality of our approach.

This paper represents a call to action to consider privacy in the design of sighted robotic systems. Privacy concerns presently prevent deployment of robotic systems in important contexts including healthcare, manufacturing and defence. We hope to have set the stage for future works to enable robotics applications currently beyond public acceptance.

%%%%%%%%%%%%%%%%%%%%%%%%%%%%%%%%%%%%%%%%%%%%%%%%%%%%%%%%%%%%%%%%%%%%%%%%%%%%%%%%

%%%References are important to the reader; therefore, each citation must be complete and correct. If at all possible, references should be commonly available publications.

\bibliographystyle{IEEEtran}
\bibliography{IEEEabrv,references}

\end{document}

%% file: MathMacros.tex
\usepackage{bm}  % bolding with \bm

\DeclareMathSymbol{\Gamma}{\mathalpha}{letters}{"00}
\DeclareMathSymbol{\Delta}{\mathalpha}{letters}{"01}
\DeclareMathSymbol{\Theta}{\mathalpha}{letters}{"02}
\DeclareMathSymbol{\Lambda}{\mathalpha}{letters}{"03}
\DeclareMathSymbol{\Xi}{\mathalpha}{letters}{"04}
\DeclareMathSymbol{\Pi}{\mathalpha}{letters}{"05}
\DeclareMathSymbol{\Sigma}{\mathalpha}{letters}{"06}
\DeclareMathSymbol{\Upsilon}{\mathalpha}{letters}{"07}
\DeclareMathSymbol{\Phi}{\mathalpha}{letters}{"08}
\DeclareMathSymbol{\Psi}{\mathalpha}{letters}{"09}
\DeclareMathSymbol{\Omega}{\mathalpha}{letters}{"0A}

\DeclareSymbolFont{EUr}{U}{eur}{m}{n}
\DeclareSymbolFont{EUb}{U}{eur}{b}{n}
\DeclareMathSymbol{\upGamma}{\mathord}{EUr}{"00}
\DeclareMathSymbol{\upDelta}{\mathord}{EUr}{"01}
\DeclareMathSymbol{\upTheta}{\mathord}{EUr}{"02}
\DeclareMathSymbol{\upLambda}{\mathord}{EUr}{"03}
\DeclareMathSymbol{\upXi}{\mathord}{EUr}{"04}
\DeclareMathSymbol{\upPi}{\mathord}{EUr}{"05}
\DeclareMathSymbol{\upSigma}{\mathord}{EUr}{"06}
\DeclareMathSymbol{\upUpsilon}{\mathord}{EUr}{"07}
\DeclareMathSymbol{\upPhi}{\mathord}{EUr}{"08}
\DeclareMathSymbol{\upPsi}{\mathord}{EUr}{"09}
\DeclareMathSymbol{\upOmega}{\mathord}{EUr}{"0A}
\DeclareMathSymbol{\upalpha}{\mathord}{EUr}{"0B}
\DeclareMathSymbol{\upbeta}{\mathord}{EUr}{"0C}
\DeclareMathSymbol{\upgamma}{\mathord}{EUr}{"0D}
\DeclareMathSymbol{\updelta}{\mathord}{EUr}{"0E}
\DeclareMathSymbol{\upepsilon}{\mathord}{EUr}{"0F}
\DeclareMathSymbol{\upzeta}{\mathord}{EUr}{"10}
\DeclareMathSymbol{\upeta}{\mathord}{EUr}{"11}
\DeclareMathSymbol{\uptheta}{\mathord}{EUr}{"12}
\DeclareMathSymbol{\upiota}{\mathord}{EUr}{"13}
\DeclareMathSymbol{\upkappa}{\mathord}{EUr}{"14}
\DeclareMathSymbol{\uplambda}{\mathord}{EUr}{"15}
\DeclareMathSymbol{\upmu}{\mathord}{EUr}{"16}
\DeclareMathSymbol{\upnu}{\mathord}{EUr}{"17}
\DeclareMathSymbol{\upxi}{\mathord}{EUr}{"18}
\DeclareMathSymbol{\uppi}{\mathord}{EUr}{"19}
\DeclareMathSymbol{\uprho}{\mathord}{EUr}{"1A}
\DeclareMathSymbol{\upsigma}{\mathord}{EUr}{"1B}
\DeclareMathSymbol{\uptau}{\mathord}{EUr}{"1C}
\DeclareMathSymbol{\upupsilon}{\mathord}{EUr}{"1D}
\DeclareMathSymbol{\upphi}{\mathord}{EUr}{"1E}
\DeclareMathSymbol{\upchi}{\mathord}{EUr}{"1F}
\DeclareMathSymbol{\uppsi}{\mathord}{EUr}{"20}
\DeclareMathSymbol{\upomega}{\mathord}{EUr}{"21}
\DeclareMathSymbol{\upvarepsilon}{\mathord}{EUr}{"22}
\DeclareMathSymbol{\upvartheta}{\mathord}{EUr}{"23}
\DeclareMathSymbol{\upvaromega}{\mathord}{EUr}{"24}
\DeclareMathSymbol{\upvarphi}{\mathord}{EUr}{"27}

\DeclareMathSymbol{\UpGamma}{\mathord}{EUb}{"00}
\DeclareMathSymbol{\UpDelta}{\mathord}{EUb}{"01}
\DeclareMathSymbol{\UpTheta}{\mathord}{EUb}{"02}
\DeclareMathSymbol{\UpLambda}{\mathord}{EUb}{"03}
\DeclareMathSymbol{\UpXi}{\mathord}{EUb}{"04}
\DeclareMathSymbol{\UpPi}{\mathord}{EUb}{"05}
\DeclareMathSymbol{\UpSigma}{\mathord}{EUb}{"06}
\DeclareMathSymbol{\UpUpsilon}{\mathord}{EUb}{"07}
\DeclareMathSymbol{\UpPhi}{\mathord}{EUb}{"08}
\DeclareMathSymbol{\UpPsi}{\mathord}{EUb}{"09}
\DeclareMathSymbol{\UpOmega}{\mathord}{EUb}{"0A}
\DeclareMathSymbol{\Upalpha}{\mathord}{EUb}{"0B}
\DeclareMathSymbol{\Upbeta}{\mathord}{EUb}{"0C}
\DeclareMathSymbol{\Upgamma}{\mathord}{EUb}{"0D}
\DeclareMathSymbol{\Updelta}{\mathord}{EUb}{"0E}
\DeclareMathSymbol{\Upepsilon}{\mathord}{EUb}{"0F}
\DeclareMathSymbol{\Upzeta}{\mathord}{EUb}{"10}
\DeclareMathSymbol{\Upeta}{\mathord}{EUb}{"11}
\DeclareMathSymbol{\Uptheta}{\mathord}{EUb}{"12}
\DeclareMathSymbol{\Upiota}{\mathord}{EUb}{"13}
\DeclareMathSymbol{\Upkappa}{\mathord}{EUb}{"14}
\DeclareMathSymbol{\Uplambda}{\mathord}{EUb}{"15}
\DeclareMathSymbol{\Upmu}{\mathord}{EUb}{"16}
\DeclareMathSymbol{\Upnu}{\mathord}{EUb}{"17}
\DeclareMathSymbol{\Upxi}{\mathord}{EUb}{"18}
\DeclareMathSymbol{\Uppi}{\mathord}{EUb}{"19}
\DeclareMathSymbol{\Uprho}{\mathord}{EUb}{"1A}
\DeclareMathSymbol{\Upsigma}{\mathord}{EUb}{"1B}
\DeclareMathSymbol{\Uptau}{\mathord}{EUb}{"1C}
\DeclareMathSymbol{\Upupsilon}{\mathord}{EUb}{"1D}
\DeclareMathSymbol{\Upphi}{\mathord}{EUb}{"1E}
\DeclareMathSymbol{\Upchi}{\mathord}{EUb}{"1F}
\DeclareMathSymbol{\Uppsi}{\mathord}{EUb}{"20}
\DeclareMathSymbol{\Upomega}{\mathord}{EUb}{"21}
\DeclareMathSymbol{\Upvarepsilon}{\mathord}{EUb}{"22}
\DeclareMathSymbol{\Upvartheta}{\mathord}{EUb}{"23}
\DeclareMathSymbol{\Upvaromega}{\mathord}{EUb}{"24}
\DeclareMathSymbol{\Upvarphi}{\mathord}{EUb}{"27}

\newcount\ppnum
\newcommand\ppnumber[1]{%
        \ppnum=#1\relax
        \ifnum\ppnum<0
                $-$%
                \ppnum=-\ppnum
        \fi
        \let\pptemp\empty
        \loop\ifnum\ppnum>999
                \count255=\ppnum
                \divide\ppnum by1000
                \count255=\numexpr \count255 - 1000*\ppnum \relax
                \edef\pptemp{,\!\ifnum\count255<100 0\ifnum\count255<10 0\fi\fi
                             \the\count255 \pptemp}%
        \repeat
        \the\ppnum
        \pptemp
}

%% file: Acronyms.tex
\newacronym{PSI}{PSI}{Photonic Systems Integration}
\newacronym{ACFR}{ACFR}{Australian Centre for Robotics}
\newacronym{CRIS}{CRIS}{Centre for Robotics and Intelligent Systems}
\newacronym{ACRA}{ACRA}{the Australasian Conference on Robotics and Automation}
\newacronym{ACRV}{ACRV}{Australian Centre for Robotic Vision}
\newacronym{USyd}{USyd}{the University of Sydney}
\newacronym{UQ}{UQ}{the University of Queensland}
\newacronym{QUT}{QUT}{the Queensland University of Technology}
\newacronym{UCSD}{UCSD}{the University of California, San Diego}
\newacronym{ANU}{ANU}{Australia National University}
\newacronym{IMOS}{IMOS}{the Integrated Marine Observation System}
\newacronym{URI}{URI}{the University of Rhode Island}
\newacronym{WHOI}{WHOI}{Woods Hole Oceanographic Institution}
\newacronym{NTNU}{NTNU}{Norwegian University of Science and Technology}

\newacronym{NSF}{NSF}{National Science Foundation}
\newacronym{LIEF}{LIEF}{Linkage Infrastructure, Equipment and Facilities}

\newacronym{ICCP}{ICCP}{the International Conference on Computational Photography}
\newacronym{CVPR}{CVPR}{Computer Vision and Pattern Recognition}
\newacronym{TIP}{TIP}{Transactions on Image Processing}
\newacronym{TSP}{TSP}{Transactions on Signal Processing}
\newacronym{JFR}{JFR}{the Journal of Field Robotics}
\newacronym{ISCAS}{ISCAS}{International Symposium on Circuits and Systems}
\newacronym{TOG}{TOG}{Transactions on Graphics}
\newacronym{ICRA}{ICRA}{International Conference on Robotics and Automation}
\newacronym{IROS}{IROS}{Intelligent Robots and Systems}
\newacronym{RA-L}{RA-L}{Robotics and Automation Letters}

\newacronym{AUV}{AUV}{autonomous underwater vehicle}
\newacronym{UAV}{UAV}{unmanned aerial vehicle}
\newacronym{USV}{USV}{unmanned surface vehicle}
\newacronym{UGV}{UGV}{unmanned ground vehicle}
\newacronym{GPS}{GPS}{global positioning system}
\newacronym{SLAM}{SLAM}{simultaneous localisation and mapping}
\newacronym{SfM}{SfM}{structure from motion}
\newacronym{AR}{AR}{augmented reality}
\newacronym{VR}{VR}{virtual reality}
\newacronym{MR}{MR}{mixed reality}
\newacronym{CNN}{CNN}{convolutional neural network}
\newacronym{DNN}{DNN}{deep neural network}
\newacronym{IMU}{IMU}{inertial measurement unit}
\newacronym{TOF}{TOF}{time of flight}

\newacronym{MDSP}{MDSP}{multi-dimensional signal processing}
\newacronym{ROS}{ROS}{region of support}
\newacronym{DOF}{DOF}{degree-of-freedom}
\newacronym{RMS}{RMS}{root mean square}
\newacronym{RMSE}{RMSE}{root mean squared error}
\newacronym{SNR}{SNR}{signal-to-noise ratio}
\newacronym{CNR}{CNR}{contrast-to-noise ratio}
\newacronym{PCA}{PCA}{principal component analysis}
\newacronym{MSE}{MSE}{mean squared error}

\newacronym{FIR}{FIR}{finite impulse response}
\newacronym{IIR}{IIR}{infinite impulse response}
\newacronym{DFT}{DFT}{discrete Fourier transform}
\newacronym{FFT}{FFT}{fast Fourier transform}
\newacronym{PSNR}{PSNR}{peak signal-to-noise ratio}
\newacronym{FPGA}{FPGA}{field programmable gate array}
\newacronym{GPU}{GPU}{graphics processing unit}
\newacronym{ASIC}{ASIC}{application-specific integrated circuit}
\newacronym{BW}{BW}{bandwidth}

\newacronym{PSF}{PSF}{point spread function}
\newacronym{SPAD}{SPAD}{single-photon avalanche diode}
\newacronym{FOV}{FOV}{field of view}
\newacronym{BRDF}{BRDF}{bidirectional reflectance distribution function}
\newacronym{FWHM}{FWHM}{full width at half maximum}
\newacronym{LF}{LF}{light field}
\newacronym{2pp}{2pp}{two-plane parameterization}
\newacronym{MLA}{MLA}{microlens array}

\newacronym{RANSAC}{RANSAC}{random sampling and consensus}
\newacronym{DoG}{DoG}{difference of Gaussian}
\newacronym{SIFT}{SIFT}{scale invariant feature transform}

\newacronym{NIR}{NIR}{near-infrared}  
\newacronym{HOG}{HOG}{histogram of oriented gradients}
\newacronym{SVM}{SVM}{support vector machine}
\newacronym{BoW}{BoW}{bag of words}

\newacronym{3D}{3D}{3-dimensional}
\newacronym{DMD}{DMD}{digital micromirror device}
\newacronym{RGB-D}{RGB-D}{red green blue-depth}